\documentclass[letterpaper]{article} % DO NOT CHANGE THIS
\usepackage[draft]{aaai2026}  % DO NOT CHANGE THIS
\usepackage{times}  % DO NOT CHANGE THIS
\usepackage{helvet}  % DO NOT CHANGE THIS
\usepackage{courier}  % DO NOT CHANGE THIS
\usepackage[hyphens]{url}  % DO NOT CHANGE THIS
\usepackage{graphicx} % DO NOT CHANGE THISa
\usepackage{natbib}  % DO NOT CHANGE THIS AND DO NOT ADD ANY OPTIONS TO IT
\usepackage{caption} % DO NOT CHANGE THIS AND DO NOT ADD ANY OPTIONS TO IT
\usepackage{algorithm}
\usepackage{algorithmic}

\usepackage{amsfonts}
\usepackage{amsmath}
\usepackage{amsthm}
\usepackage{amssymb}
\usepackage{booktabs}
\usepackage{mathtools}
\usepackage{multirow}
\usepackage{subcaption}
\usepackage{xcolor}

\newtheorem{theorem}{Theorem}[section]
\theoremstyle{definition}
\newtheorem{definition}[theorem]{Definition}
\newtheorem{proposition}[theorem]{Proposition}

\usepackage{newfloat}
\usepackage{listings}
\DeclareCaptionStyle{ruled}{labelfont=normalfont,labelsep=colon,strut=off} % DO NOT CHANGE THIS
\floatstyle{ruled}
\newfloat{listing}{tb}{lst}{}
\floatname{listing}{Listing}
\title{ Prediction Loss Guided Decision-Focused Learning }
\author{
    Haeun Jeon\textsuperscript{\rm 1},
    Hyunglip Bae\textsuperscript{\rm 1},
    Chanyeong Kim\textsuperscript{\rm 1},
    Yongjae Lee\textsuperscript{\rm 2}\corresponding,
    Woo Chang Kim\textsuperscript{\rm 1}\corresponding
}
\affiliations{
    \textsuperscript{\rm 1}KAIST\\
    \textsuperscript{\rm 2}UNIST\\
    haeun39@kaist.ac.kr, qogudflq@kaist.ac.kr, kim.chanyeong@kaist.ac.kr, yongjaelee@unist.ac.kr, wkim@kaist.ac.kr
}

\begin{document}

\maketitle

\begin{abstract}
Decision‑making under uncertainty is often considered in two stages: predicting the unknown parameters, and then optimizing decisions based on predictions.
While traditional prediction-focused learning (PFL) treats these two stages separately, decision-focused learning (DFL) trains the predictive model by directly optimizing the decision quality in an end-to-end manner.
However, despite using exact or well-approximated gradients, vanilla DFL often suffers from unstable convergence due to its flat-and-sharp loss landscapes.
In contrast, PFL yields more stable optimization, but overlooks the downstream decision quality.
To address this, we propose a simple yet effective approach: perturbing the decision loss gradient using the prediction loss gradient to construct an update direction.
Our method requires no additional training and can be integrated with any DFL solvers.
Using the sigmoid-like decaying parameter, we let the prediction loss gradient guide the decision loss gradient to train a predictive model that optimizes decision quality.
Also, we provide a theoretical convergence guarantee to Pareto stationary point under mild assumptions.
Empirically, we demonstrate our method across three stochastic optimization problems, showing promising results compared to other baselines. 
We validate that our approach achieves lower regret with more stable training, even in situations where either PFL or DFL struggles.
\end{abstract}


%%%%%%%%%%%%%%%%%%%%%%%%%%%%%%%%%%%%%%%%%% SECTION: INTRO %%%%%%%%%%%%%%%%%%%%%%%%%%%%%%%%%%%%%%%%%%
\section{Introduction}
\label{sec:intro}
%%%%%%%%%%%%%%%%%%%%% NOTES %%%%%%%%%%%%%%%%%%%%%
% \begin{itemize}
%     \item contribution summary
%     \item can be used with any differentiable optimizer
%     \item cost efficient (need to emphasize)
%     \item need figure of 
%     \item we are the first to perturb gradients of exact dfl methods
%     \item note that we are not surrogate loss methods
% \end{itemize}

We study decision-making under uncertainty through the model
\begin{equation}
a^*(y) \;=\;\arg\min_{a\in\mathcal{A}}\,f(a,y)
\quad\text{with}\quad
y \;=\; M_\theta(x),
\end{equation}
with decision variable $a$, uncertain parameter $y$, known feasible set $\mathcal{A}$, objective function $f$, and a predictive model $M_{\theta}$ parameterized by $\theta$ that maps input features $x$ to an estimate of $y$.
This two‐stage formulation underlies a variety of applications such as vehicle routing, portfolio optimization, inventory control, and the knapsack problem, where one first forecasts the uncertain parameters and then solves for the optimal decision \citep{mandi2020smart, elmachtoub2022smart, donti2017task, wilder2019melding}.

Traditionally, such problems have been addressed via the predict-focused learning (PFL) paradigm.
A predictive model in PFL is trained solely to minimize the error between $M_\theta(x)$ and the true parameter $y$.
Once trained, these forecasts are fed into a downstream optimizer to select $a^*$.
The underlying belief in PFL is that good predictions will yield better decisions; yet even small forecast errors can cascade into horrible decisions.
Consequently, in practice, prediction and optimization are deeply entwined and cannot always be decoupled without loss of performance.

To bridge this gap, decision-focused learning (DFL) has been developed.
DFL embeds the optimization layer into the training loop by defining a \textit{decision loss} that measures the quality of decisions.
Then, the decision loss is backpropagated through the solver.
This backpropagation requires special techniques such as implicit differentiation via KKT conditions \citep{amos2017optnet, donti2017task}, perturbation-based gradient estimators \citep{wilder2019melding}, or fully differentiable convex optimization layers \citep{agrawal2019differentiable} to obtain gradients for end-to-end training.

Despite leveraging exact or well-approximated gradients, vanilla DFL often shows unstable convergence behavior.
Its decision loss landscape exhibits abrupt jumps and flat regions \citep{bansal2024taskmet}.
To mitigate these geometric hurdles, researchers have introduced smooth surrogate losses and objectives that blend prediction and decision losses, aiming for stable training.
However, these existing approaches suffer from three key limitations.
First, many are tailored to a narrow class of optimization problems and cannot be easily generalized to other decision-making settings.
Second, learning the surrogate loss models often incurs substantial computational overhead, undermining their practicality for large-scale or time-sensitive applications.
Third, by relying on a fixed weighting scheme to combine prediction and task losses, they lack the flexibility to dynamically adjust the balance between prediction and decision loss as training progresses.

In this paper, we begin with a geometric analysis showing discrepancies in both direction and magnitude between PFL and DFL gradients, caused by their distinct loss landscape structures.
Building on this insight, we propose \textit{Prediction Loss Guided Decision-Focused Learning}, a gradient-perturbation framework that injects prediction loss gradients into the decision loss update.
We compute a new descent direction by bisecting the angle between the two gradient vectors and then shifting toward the decision loss gradient under a tunable parameter $\kappa$.
We prove that our model with $\kappa=0$ converges to a Pareto stationary point.
Finally, through experiments on three stochastic optimization tasks, we demonstrate that our method outperforms baselines in minimizing decision loss.

Overall, this paper makes the following contributions:
\begin{itemize}
    \item We avoid constructing any surrogate loss models. Our method directly perturbs the original gradients, reducing the computational overhead compared to existing surrogate loss approaches.
    \item Our method can be combined with any vanilla DFL algorithms, making it universally applicable across diverse frameworks.
    \item To our knowledge, this is the first study to explicitly adjust the DFL gradient direction (rather than tweaking the loss) to boost decision-making performance.
    \item We provide the in-depth geometric comparison that quantifies the directional and magnitude discrepancies between prediction and decision loss gradients.
    \item We prove that our method converges to a Pareto stationary point under mild assumptions.
    \item We conduct experiments on three stochastic optimization problems and demonstrate that our method outperforms other baselines.
\end{itemize}

%%%%%%%%%%%%%%%%%%%%%%%%%%%%%%%%%%%%%%%%%% SECTION: RELATED WORKS %%%%%%%%%%%%%%%%%%%%%%%%%%%%%%%%%%%%%%%%%%
\section{Related Works}
\label{sec:related}

\paragraph{Decision-Focused Learning}
Gradient-based DFL has been explored through several complementary strategies \cite{mandi2023decision}. One line of work directly differentiates through the constrained optimization layer: convex quadratic programs have been handled by applying KKT-based backpropagation with OptNet for efficiency \cite{amos2017optnet,donti2017task}, and this idea was generalized to arbitrary convex problems via the differentiable solver Cvxpylayers \cite{agrawal2019differentiable}.

A second family of methods smooths the mapping from predicted parameters to decisions by adding regularizer.
Examples include appending the Euclidean norm of decision variables to enable quadratic programming based differentiation \cite{wilder2019melding}, incorporating a logarithmic barrier to make linear programs differentiable \cite{mandi2010interior}, and using entropy penalties to tackle multi‐label tasks \cite{martins2017learning,amos2019limited}.
Perturb-and-MAP frameworks further regularize via random perturbations to the objective or constraints \cite{papandreou2011perturb,niepert2021implicit,berthet2020learning}.

More recently, researchers have designed explicit surrogate losses to approximate the true task loss in a differentiable way.
The SPO+ loss \cite{elmachtoub2022smart} provides a convex upper bound with computable subgradients; NCE \cite{mulamba2020contrastive} and its ranking extension LTR \cite{mandi2022decision} derive loss families via noise‐contrastive estimation; SO-EBM \cite{kong2022end} employs an energy‐based layer as a proxy optimizer; and LODL \cite{shah2022decision} builds local parametric surrogates (e.g.\ WeightedMSE, Quadratic) around each instance, later extended to a global model in EGL \cite{shah2024leaving} and LANCER \cite{zharmagambetov2024landscape}.
Most recently, a Lipschitz‐continuous perturbation gradient loss has been proposed whose approximation error vanishes with increasing samples \cite{huang2024decision}.

Unlike these prior methods, our work dynamically blends the prediction and decision loss gradients to directly enhance DFL performance without constructing surrogate models.

\paragraph{Gradient Perturbing Methods}
Since our method treats the prediction loss as a guide to steer the optimization of the decision loss, it can be interpreted as a form of multi-task learning.
We therefore explore three canonical multi-task algorithms: projecting conflicting gradients (PCGrad) \citep{yu2020gradient}, multiple gradient descent algorithm (MGDA) \citep{desideri2012multiple}, and dual cone gradient descent (DCGD) \citep{hwang2024dual}.

PCGrad is an algorithm designed to tackle conflicting gradients.
At each update, PCGrad computes the gradient for each task.
Whenever two gradients conflict (having negative cosine similarity), PCGrad projects one gradient onto the normal plane of the other.
The projection removes the component that can degrade the performance of other tasks.
This gradient surgery method guarantees not to increase other task losses while updating.
MGDA tackles multi‐task learning by identifying a single update direction that decreases all objectives simultaneously.
It computes the gradient by finding the minimum norm descent step under the gradients' convex hull.
This guarantees no increase in every loss.
A suitable step size is chosen that strictly decreases all objectives.
DCGD is a technique that ensures updates simultaneously decrease all component losses by constraining the descent direction to lie within the dual cone of the individual gradient vectors.
At each iteration, DCGD characterizes the cone generated by the gradients of each task's losses.
Then, it computes its dual cone (the set of directions having non‐negative inner products with both gradients), and then selects a feasible update to reduce each loss term.

We utilize these three multi-task learning algorithms as baselines and benchmark their performance against our proposed method.

%%%%%%%%%%%%%%%%%%%%% FIG: FAKE %%%%%%%%%%%%%%%%%%%%%
\begin{figure}[t]
% \vskip -0.1in
\begin{center}
\centerline{\includegraphics[width=0.48\textwidth]{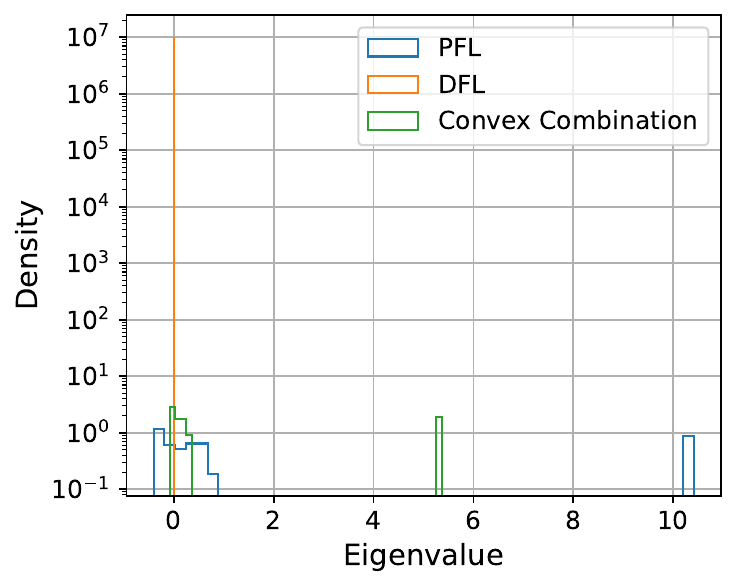}}
\caption{
Hessian eigenvalue density plot of the three different losses $\mathcal{L}_{pred}$, $\mathcal{L}_{dec}$, and $(1-\beta)\mathcal{L}_{pred} + \beta \mathcal{L}_{dec}$ with $\beta=0.5$ with respect to the predictive model parameters.
The eigenvalues are evaluated using the Lanczos algorithm at epoch 2 in the portfolio optimization problem.
The eigenvalues reflect the curvature of the loss surface along different directions: large eigenvalues correspond to steep regions, while values near zero indicate flatness.
The spectrum of $\mathcal{L}_{dec}$ is concentrated near zero, suggesting that the decision loss landscape is extremely flat.
In contrast, $\mathcal{L}_{pred}$ shows saddle-like curvature, as indicated by a broader spread including negative eigenvalues and few large ones.
The convex combination yields a more balanced distribution, blending the flatness of decision loss with the curvature of prediction loss.
This suggests that even a simple equal-weighted convex combination can enrich the gradient signal for training, allowing $\mathcal{L}_{pred}$ to effectively guide the optimization of $\mathcal{L}_{dec}$.
}
\label{fig:hessian}
\end{center}
% \vskip -0.3in
\end{figure}

%%%%%%%%%%%%%%%%%%%%% FIG: MAG_COS %%%%%%%%%%%%%%%%%%%%%
\begin{figure*}[t]
\centering
\begin{subfigure}{.35\textwidth}
  \centering
  \includegraphics[width=\textwidth]{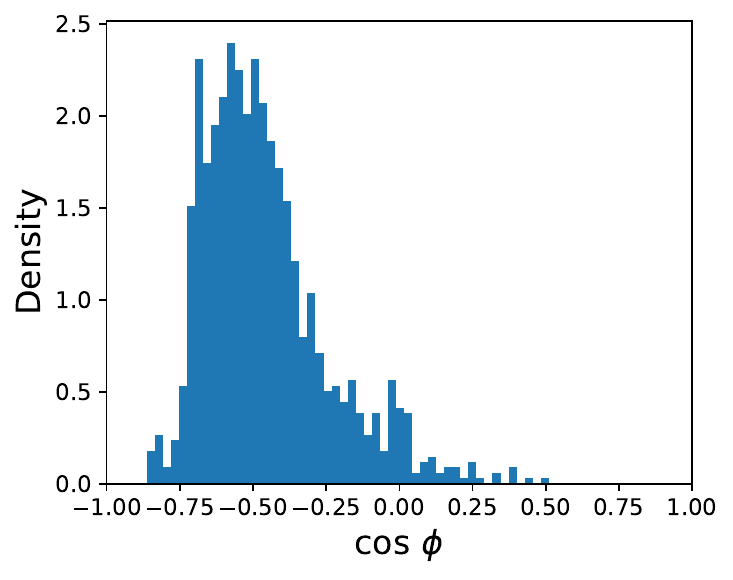}
  \caption{Histogram of cosine similarity}
  \label{subfig:cos}
\end{subfigure}%
\begin{subfigure}{.35\textwidth}
  \centering
  \includegraphics[width=\textwidth]{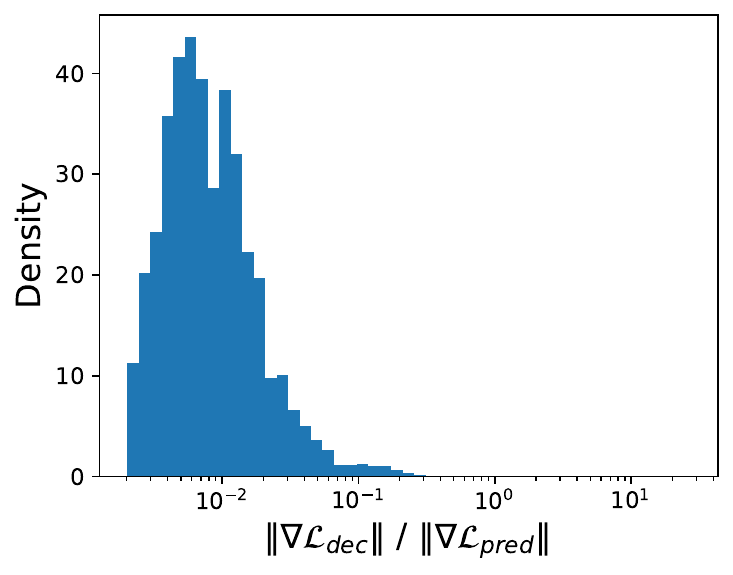}
  \caption{Histogram of magnitude ratio}
  \label{subfig:mag}
\end{subfigure}
\begin{subfigure}{.29\textwidth}
  \centering
  \includegraphics[width=\textwidth]{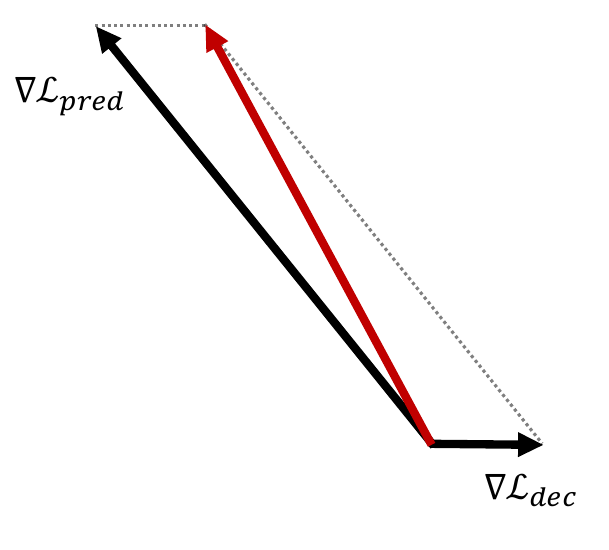}
  \caption{An illustrative example}
  \label{subfig:magcos-grad}
\end{subfigure}
\caption{
Comparison of prediction loss gradients $\nabla \mathcal{L}_{pred}$ and decision loss gradients $\nabla \mathcal{L}_{dec}$ during training on the budget allocation problem.
Each subplot represents:
\textbf{(a)} the cosine similarity between $\nabla \mathcal{L}_{pred}$ and $\nabla \mathcal{L}_{dec}$,
\textbf{(b)} the gradient norm ratio $ \| \nabla \mathcal{L}_{dec} \| / \| \nabla \mathcal{L}_{pred} \|$ on a log scale, and
\textbf{(c)} an illustrative case where two gradients with characteristics of (a) and (b) are naively added.
In subplot (a), cosine similarity values near $1$ indicate that the gradients are aligned in the same direction, while values near $–1$ indicate they point in opposite directions.
The figure shows that in most cases, $\cos\phi < 0$ for budget allocation, meaning the gradients are often conflicting.
Subplot (b) shows that in many instances, the magnitude of $\nabla \mathcal{L}_{dec}$ is approximately 100 times smaller than that of $\nabla \mathcal{L}_{pred}$.
Subplot (c) demonstrates that naively summing two conflicting and imbalanced may lead to biased training.
}
\label{fig:mag-cos}
\end{figure*}

%%%%%%%%%%%%%%%%%%%%%%%%%%%%%%%%%%%%%%%%%% SECTION: Prediction Guided DFL %%%%%%%%%%%%%%%%%%%%%%%%%%%%%%%%%%%%%%%%%%
% \section{Phenomenon on Two Gradients}
\section{Methodology}
\label{sec:meth}

While many differentiable decision-focused learning (DFL) methods have been developed, little attention has been paid to how these methods behave under the complex geometry of the decision loss landscape.
In this section, we formally define the DFL problem setting with notations and examine the distinct properties of gradients of prediction and decision losses in DFL.
We then introduce our proposed approach to mitigate the empirically observed challenges and provide theoretical analyses of our method.

%%%%%%%%%%%%%%%%%%%%% subsec: problem settings %%%%%%%%%%%%%%%%%%%%%
\subsection{Problem Settings}

We begin by formulating the problem in two stages: the \textit{prediction stage} and the \textit{optimization stage}.
The prediction stage involves estimating unknown parameters for the downstream task.
The optimization stage uses these predictions to compute the optimal decisions.

Given a input features $x$ and their ground-truth $y$, our objective is to train a predictive model $\mathcal{M}_{\theta}$ parameterized by $\theta$, to output predictions $\hat{y} = \mathcal{M}_{\theta}(x)$.
These predictions are used to compute the optimal decisions (or actions) $a^*(\hat{y})$ in the optimization stage that minimizes the decision loss $\mathcal{L}_{dec} (a^*(\hat{y}), y)$.
We assume that the unknown parameters are only in the objective function, not in the constraints of the downstream optimization problem.

Prediction-focused learning (PFL) trains the predictive model $\mathcal{M}_{\theta}$ to minimize the discrepancy between the prediction $\hat{y}$ and the ground truth $y$, typically via a standard loss function such as mean-squared error (MSE).
% such as:
% \begin{equation}
%     \min_{\theta} \| \mathcal{M}_{\theta}(x) - y \|
% \end{equation}
% where $\| \cdot \|$ denotes the Euclidean norm.
The gradient with respect to the predictive model parameters $\theta$ is computed as:
\begin{equation}
    \nabla_{\theta} \mathcal{L}_{pred} (\hat{y}, y)
    = \frac{ \partial \mathcal{L}_{pred}( \hat{y}, y) }{ \partial \hat{y} }
    \cdot
    \frac{ \partial \hat{y} }{ \partial \theta } \\
\end{equation}

In the training stage of PFL, the downstream optimization stage is ignored, based on the underlying belief that accurate predictions will lead to high-quality decisions.
After the predictive model $\mathcal{M}_{\theta}$ is trained, PFL uses the learned $\mathcal{M}_{\theta}$ to derive predictions, which are then passed to the solver to compute the optimal decision $a^*(\hat{y})$.

Instead, DFL treats the prediction stage and the optimization stage in an end-to-end manner.
DFL updates $\mathcal{M}_{\theta}$ using the gradients that directly optimize the decision loss:
\begin{align}
    \nabla_{\theta} \mathcal{L}_{dec} (a^*(\hat{y}), y)
    % &= \frac{ \partial \mathcal{L}_{dec}( a^*(\hat{y}), y) }{ \partial a^*(\hat{y}) }
    % \cdot
    % \frac{ \partial a^*(\hat{y}) }{ \partial \theta } \\
    &= \frac{ \partial \mathcal{L}_{dec}( a^*(\hat{y}), y) }{ \partial a^*(\hat{y}) }
    \cdot
    \frac{ \partial a^*(\hat{y}) }{ \partial \hat{y} }
    \cdot
    \frac{ \partial \hat{y} }{ \partial \theta}
\end{align}

The primary challenge lies in computing $\frac{ \partial a^*(\hat{y}) }{ \partial \hat{y} }$, which requires differentiating through the optimization solver.
Prior work has addressed this challenge using a variety of techniques, including implicit differentiation via the Karush-Kuhn-Tucker (KKT) conditions \citep{amos2017optnet, donti2017task}, perturbation-based approximations \citep{wilder2019melding}, and differentiable convex optimization layers \citep{agrawal2019differentiable}.

%%%%%%%%%%%%%%%%%%%%% SUBSECTION: Lp and Ld Observation %%%%%%%%%%%%%%%%%%%%%
\subsection{Empirical Observation on $\nabla \mathcal{L}_{pred}$ and $\nabla \mathcal{L}_{dec}$}
% \begin{itemize}
%     \item want to dfl but mapping is flat and sharp (show hessian)
%     \item want to use pfl as a guide; somewhat like highway and local road
%     \item previous research used pfl+dfl or pfl warm-start
%     \item however pfl dfl gradient cosine similarity is different (show cos)
% \end{itemize}

To understand the challenges in DFL, we begin by analyzing the behavior of the prediction loss gradients $\nabla \mathcal{L}_{pred}$ and the decision loss gradients $\nabla \mathcal{L}_{dec}$.
Although both gradients ultimately head towards the equivalent global optimal where the prediction $\hat{y}$ is equal to the ground truth $y$, they differ significantly in direction and magnitude due to the geometry of their respective loss landscapes.
These differences can cause instability when the gradients are naively combined.

We first investigate the curvature of the decision loss landscape with respect to the model’s predictions.
Unlike prediction losses, which are typically smooth and convex, the decision loss surface is often non-convex and exhibits both flat regions and sharp minima.
We illustrate this in Figure \ref{fig:hessian} by computing Hessian eigenvalue density of three cases: $\mathcal{L}_{pred}$, $\mathcal{L}_{dec}$ and $\mathcal{L}_{pred} + \mathcal{L}_{dec}$ in the portfolio optimization problem.
In the figure, the Hessian eigenvalues of $\mathcal{L}_{dec}$ are mostly near-zero values, indicating the flat loss landscape.
Instead, when the objective is augmented with $\mathcal{L}_{pred}$, the landscape becomes significantly smoother.

Next, we further analyze the directional and magnitude differences between $\mathcal{L}_{pred}$ and $\mathcal{L}_{dec}$ in Figure \ref{fig:mag-cos}.
To quantify directionality, we compute the cosine similarity $\cos \phi$ between the two gradients, where $\phi$ is the angle between them.
Two gradients form an obtuse angle when $-1 \leq \cos \phi < 0$ and an acute angle when $0 \leq \cos \phi < 1$.
We also compute the ratio of their norms as $\| \mathcal{L}_{dec} \| / \| \mathcal{L}_{pred} \|$ to assess relative scale.
Figure \ref{subfig:cos} shows that the cosine similarity is often negative, indicating frequent directional conflict.
Moreover, as shown in Figure \ref{subfig:mag}, $\| \nabla \mathcal{L}_{pred} \|$ is typically 10 to 1000 times larger than $\| \nabla \mathcal{L}_{dec} \|$.
This imbalance suggests that the influence of $\nabla \mathcal{L}_{dec}$ can be easily overshadowed in naive convex combinations settings.
We visualize an example with negative cosine similarity with a tiny magnitude ratio in Figure \ref{subfig:magcos-grad}, where the combined gradient (red) is dominated by $\nabla \mathcal{L}_{pred}$.

These empirical findings highlight the need for a more structured approach to gradient combination; one that balances direction and scale while preserving the influence of both loss components.

%%%%%%%%%%%%%%%%%%%%% FIG: OURS %%%%%%%%%%%%%%%%%%%%%
\begin{figure*}[t]
\centering
% \vspace{-30pt}
\includegraphics[width=\textwidth]{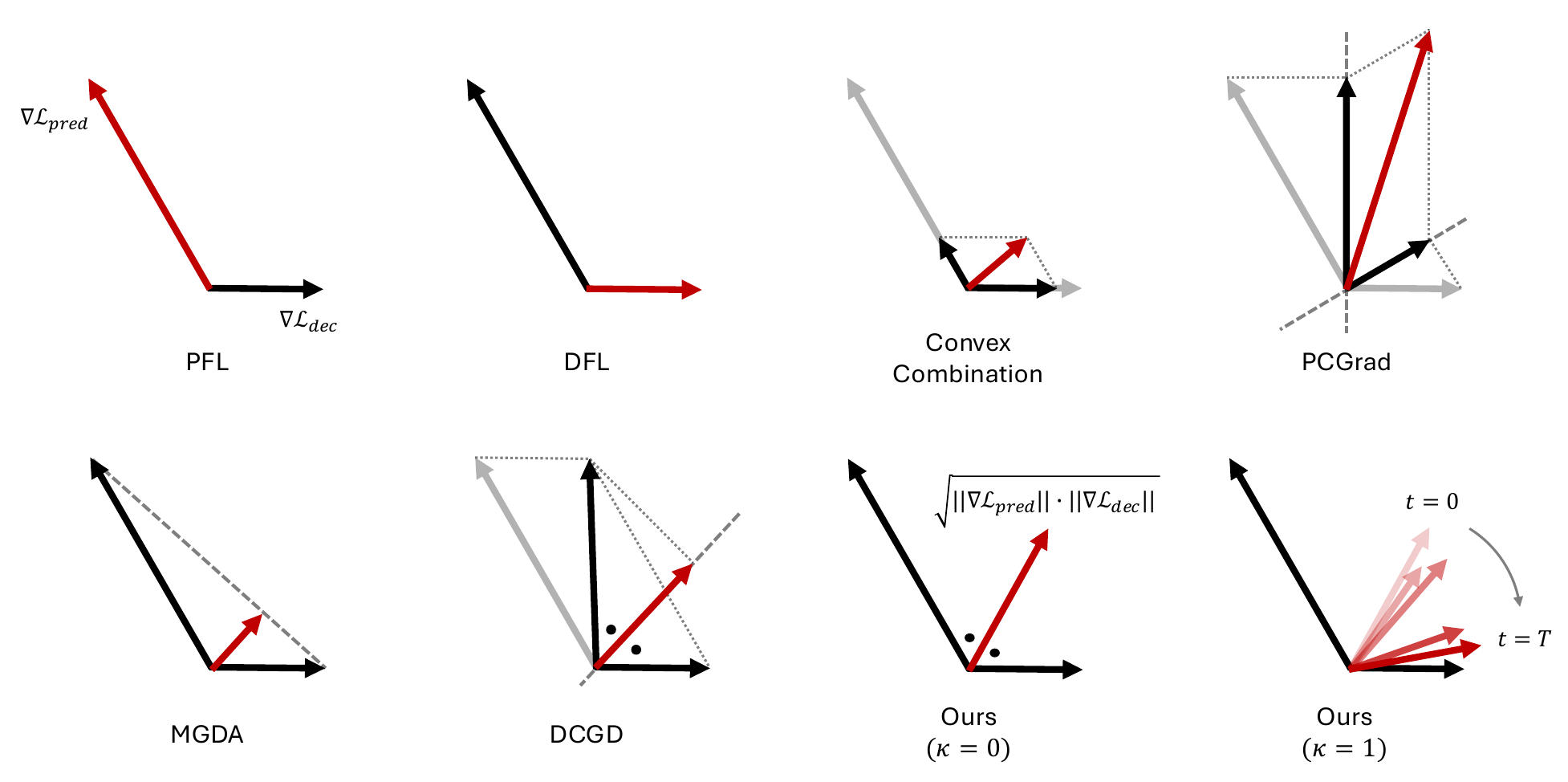}
\caption{
This figure illustrates the update directions used by our method and the baselines.
From top-left to bottom-right, the approaches shown are: PFL, DFL, Convex Combination, PCGrad, MGDA, DCGD, our method with $\kappa=0$, and $\kappa=1$.
$ \nabla \mathcal{L}_{pred}$ and $ \nabla \mathcal{L}_{dec}$ are identical across all baselines for easy comparison.
The red arrow indicates the gradient used for model updates.
PFL and DFL use only their respective gradients for updates.
The convex combination method optimizes the loss $(1 - \beta)\mathcal{L}_{pred} + \beta\mathcal{L}_{dec}$ with $\beta \in [0,1]$, combining both gradients directly.
PCGrad projects each gradient onto the normal space of the other before summing if two gradients conflict.
MGDA finds the minimum-norm direction within the convex hull of the two gradients, while DCGD merges the gradients and projects the result onto the bisecting vector defined by the merged vector and $\nabla \mathcal{L}_{dec}$.
Our method is visualized in two variants.
When $\kappa = 0$, we use the gradient that bisects $\nabla \mathcal{L}_{pred}$ and $\nabla \mathcal{L}_{dec}$ for the update.
The gradient norm is computed as the geometric mean of $\| \nabla \mathcal{L}_{pred} \|$ and $\| \nabla \mathcal{L}_{dec} \|$.
When $\kappa = 1$, a decaying weight is applied to $\nabla \mathcal{L}_{pred}$, gradually shifting the update direction toward $\nabla \mathcal{L}_{dec}$ as training progresses from epoch $t=0$ to $T$.
Here, lighter red arrows indicate earlier epochs, with darker arrows showing later stages of training.
We depict varying arrow lengths to reflect changes in gradient norms as the training progresses.
}
% \vskip -0.3in
\label{fig:gradients}
\end{figure*}

%%%%%%%%%%%%%%%%%%%%% SUBSECTION: METHODOLOGY %%%%%%%%%%%%%%%%%%%%%
\subsection{Prediction Loss Guided Decision-Focused Learning}
% \begin{itemize}
%     \item we use center method to minimize dec loss
%     \item our method minimizes not only the dec loss, but reduces the pred loss somewhat like Pareto stationary point
% \end{itemize}

We now introduce our gradient perturbation framework for DFL.
Motivated by the empirical findings in the previous section, our method aims to mitigate the optimization instability caused by the flat and sharp decision loss landscape.
In particular, we guide the direction of $\nabla \mathcal{L}_{dec}$ using $\nabla \mathcal{L}_{pred}$, while adaptively balancing the influence of each.

In the early stages of training, the predictive model $\mathcal{M}_{\theta}$ is typically uncalibrated and may produce inaccurate or unstable outputs.
At this point, relying solely on $\nabla \mathcal{L}_{dec}$, which are often noisy, low in magnitude, and influenced by its flat and sharp loss landscape, can lead to poor updates.

To address this, we initially aim to reduce both $\mathcal{L}_{pred}$ and $\mathcal{L}_{dec}$ simultaneously.
This ensures that the model begins by learning to approximate the unknown parameters, providing a more stable input for the downstream optimization task.

Let $u_{pred}$ and $u_{dec}$ denote the unit norm direction of $\nabla \mathcal{L}_{pred}$ and $\nabla \mathcal{L}_{dec}$.
% \begin{equation}
%     u_{i} \vcentcolon=  \frac{ \nabla \mathcal{L}_{i} } { \| \nabla \mathcal{L}_{i} \| }, \quad i \in \{pred, dec\}. \\
% \end{equation}
First, we choose the gradient that bisects $u_{pred}$ and $u_{dec}$.
This direction seeks to improve both objectives.
Then, as training progresses, we gradually decrease the influence of $\nabla \mathcal{L}_{pred}$, allowing the update direction to converge toward $\nabla \mathcal{L}_{dec}$.
This shifting reflects a central insight of our approach: while prediction accuracy is helpful in early training, decision quality is the ultimate target.
Our method implements this transition in a smooth, principled manner using a decay parameter $\alpha$ controlled by an inflection point $c$ and steepness parameter $\kappa$:
\begin{equation}
    \alpha \vcentcolon= \bigl(1 + e^{t - c}\bigr)^{-\kappa}
\end{equation}
where $t$ is the training epoch.
Note that $\alpha$ is a monotonically decreasing function of $t$ for $\kappa \geq 0$ satisfying $0 < \alpha \leq 1$.

To incorporate the gradient norm information, we compute the geometric mean $m$ of the two gradient norms:
\begin{equation}
    m \vcentcolon= \sqrt{\| \nabla \mathcal{L}_{pred} \| \cdot \| \nabla \mathcal{L}_{dec} \|}
\end{equation}

Finally, we define the merged gradient $g$ for the update as:
\begin{equation} \label{eq:g}
    g \vcentcolon= m \cdot \frac{ \alpha \cdot u_{pred} + u_{dec}}{\| \alpha \cdot u_{pred} + u_{dec} \|} \\
\end{equation}
    
In practice, we fix the inflection parameter $c$ and consider two settings: $\kappa=0$ and $\kappa=1$.
When $\kappa=0$, $\alpha=1$ throughout training, and therefore the updating gradient $g$ always points the direction that bisects the angle between $\nabla \mathcal{L}_{pred}$ and $\nabla \mathcal{L}_{dec}$.
When $\kappa = 1$, $\alpha$ gradually decreases, allowing the update to lean more heavily toward $\nabla \mathcal{L}_{dec}$ as training progresses.
Intuitive update directions for both models are illustrated in Figure \ref{fig:gradients}.
We compare the empirical performance of these two variants in Section \ref{sec:exp}.

%%%%%%%%%%%%%%%%%%%%% SUBSECTION: Convergence Analysis %%%%%%%%%%%%%%%%%%%%%
\subsection{Theoretical Analysis}
In this section, we present the theoretical analysis of our proposed method.
We begin by defining what it means for two gradients to conflict.
Using the definition, we show that our update rule has conflict-avoidance properties.
Then, we introduce the combined loss and Pareto concepts.
Finally, we prove that under mild assumptions, our method converges to a Pareto stationary point when $\kappa=0$.
All proofs can be found in Appendix \ref{appen:proof}.

First, we define the gradient conflict in terms of the angle between two vectors and present a proposition demonstrating that our method avoids gradient conflict.
%%%%%%%%%%%%%%%%%%%%% DEF: CONFLICTING GRAD %%%%%%%%%%%%%%%%%%%%%
\begin{definition}[Conflicting Gradients] \label{def:conflict}
Let $\phi_{i,j}$ denote the angle between two vectors $v_i$ and $v_j$.
We say that two gradeints \textit{conflict} if $\cos{\phi_{i,j}} < 0$.
\end{definition}

%%%%%%%%%%%%%%%%%%%%% THM: conflicting %%%%%%%%%%%%%%%%%%%%%
\begin{proposition}[No Confliction]
\label{prop:no-conf}
Our method never conflicts with $\nabla \mathcal{L}_{dec}$.
Moreover, when $\kappa=0$, our method does not conflict with both $\nabla \mathcal{L}_{pred}$ and $\nabla \mathcal{L}_{dec}$.
\end{proposition}

Proposition \ref{prop:no-conf} states that the proposed method always updates in a direction that does not conflict with $\nabla \mathcal{L}_{dec}$.
In practice, as shown in Figure \ref{fig:mag-cos}, $\nabla \mathcal{L}_{pred}$ and $\nabla \mathcal{L}_{dec}$ frequently conflict, and $\| \nabla \mathcal{L}_{dec} \|$ tends to be much smaller than $\| \nabla \mathcal{L}_{pred} \|$.
In such cases, naive gradient combinations can result in an update direction that conflicts with $\nabla \mathcal{L}_{dec}$ as illustrated in Figure \ref{subfig:magcos-grad}.
Our method avoids this issue by ensuring that the update direction never conflicts with $\nabla \mathcal{L}_{dec}$ regardless of the directions of two gradients or their relative magnitudes.

Now, we define the Pareto concepts that can be used for convergence analysis.
% Now, we define the combined loss $\mathcal{L}$ and the Pareto concepts that can be used for convergence analysis.
% %%%%%%%%%%%%%%%%%%%%% DEF: combined loss %%%%%%%%%%%%%%%%%%%%%
% \begin{definition}[Combined Loss]   \label{def:combined-loss}
% Consider the prediction loss $\mathcal{L}_{pred}$ and the decision loss $\mathcal{L}_{dec}$.
% We define the combined loss as $\mathcal{L} = \mathcal{L}_{pred} + \mathcal{L}_{dec}$.
% \end{definition}

%%%%%%%%%%%%%%%%%%%%% DEF: Pareto %%%%%%%%%%%%%%%%%%%%%
\begin{definition}[Pareto optimal and stationary points]
For the prediction loss $\mathcal{L}_{pred}$ and decision loss $\mathcal{L}_{dec}$, a point $w^*$ is Pareto optimal if there is no other $w$ such that $     \mathcal{L}_{pred}(w) \leq \mathcal{L}_{pred}(w^*)$ and $\mathcal{L}_{dec}(w) \leq \mathcal{L}_{dec}(w^*)$.
% \begin{gather}
%     \mathcal{L}_{pred}(w) \leq \mathcal{L}_{pred}(w^*) \\
%     \mathcal{L}_{dec}(w) \leq \mathcal{L}_{dec}(w^*)
% \end{gather}
A point $w^o$ is Pareto stationary if there exists $\alpha_1, \alpha_2 \geq 0$ with $\alpha_1 + \alpha_2 = 1$ such that $ \alpha_1 \nabla \mathcal{L}_{pred}(w^o) + \alpha_2 \nabla \mathcal{L}_{dec}(w^o) = \textbf{0} $,
% \begin{equation}
%     \alpha_1 \nabla \mathcal{L}_{pred}(w^o) + \alpha_2 \nabla \mathcal{L}_{dec}(w^o) = \textbf{0}
% \end{equation}
i.e. there exists a zero vector in the convex hull of two gradients.
\end{definition}

The definition of a Pareto stationary point indicates that no small perturbation in any direction can simultaneously reduce both the prediction and decision losses.
Finally, we show that our method converges to a Pareto-stationary point when $\kappa=0$.

%%%%%%%%%%%%%%%%%%%%% THM: convergence %%%%%%%%%%%%%%%%%%%%%
\begin{theorem}[Convergence to Pareto Stationary Point] \label{thm:pareto}
Assume $\mathcal{L}_{pred}$ and  $\mathcal{L}_{dec}$ are differentiable with $M_i$-Lipschitz continuous with $M_i>0$.
Then, our method with $\kappa=0$ converges to a Pareto-stationary point for the step size $\eta_k$ satisfying $ \sum_{k=0}^\infty\eta_k=\infty$ and $\sum_{k=0}^\infty\eta_k^2<\infty$.
% \begin{gather}
%     \sum_{k=0}^\infty\eta_k=\infty,
%     \quad
%     \sum_{k=0}^\infty\eta_k^2<\infty
% \end{gather}
Moreover, the algorithm converges at a rate upper bounded by $\mathcal{O}(1/\sqrt{T})$.
\end{theorem}

Theorem \ref{thm:pareto} shows that our method with $\kappa=0$ converges to a Pareto stationary point at a rate of $\mathcal{O}(1/\sqrt{T})$.
This theoretical guarantee aligns with previous studies established for gradient perturbing methods such as PCGrad, MGDA, and DCGD, which are introduced in Section \ref{sec:related}.

% Although the primary goal in DFL is to minimize $\mathcal{L}_{dec}$ rather than jointly minimizing $\mathcal{L}_{pred}$ and $\mathcal{L}_{dec}$, we claim that if $\mathcal{L}_{dec}$ is well minimized empirically, Theorem \ref{thm:pareto} gives an additional theoretical guarantee that we can not improve both losses from that point.

%%%%%%%%%%%%%%%%%%%%% TABLE: Results %%%%%%%%%%%%%%%%%%%%%
\begin{table*}[t]
    % \vskip 0.1in
    \centering
    \caption{
    Results for five experimental settings evaluated with normalized test regret and its standard error of the mean (SEM).
    The metric is lower the better and 0 when optimal.
    We report separate results for the weighted and unweighted variants of the knapsack problem, as well as for the budget allocation problem with 0 and 500 fake targets.
    The lowest regret values for each problem are bold-lettered.
    Our method outperforms the baselines across the problems except for the unweighted knapsack.
    The SEM of our method is consistently low, indicating reliable performance across all domains.
    }
    \begin{small}
    \begin{sc}
    % \setstretch{1.3}
    \renewcommand{\arraystretch}{1.3}
        \begin{tabular}{crccccc}
        \toprule
        \multicolumn{2}{c}{\multirow{2}{*}{\textbf{Methods}}} & \multicolumn{5}{c}{\textbf{Problems}} \\
        \cline{3-7}
        &                              & Knapsack(UW) & Knapsack(W) & Budget(0) & Budget(500)  & Portfolio  \\
        \midrule
        \multirow{2}{*}{Naive}
        & \texttt{PFL}                 & 0.258 $\pm$ 0.318 & 0.360 $\pm$ 0.265 & 0.425 $\pm$ 0.236 & 0.577 $\pm$ 0.044 & 0.515 $\pm$ 0.012  \\
        & \texttt{DFL}                 & 0.190 $\pm$ 0.301 & 0.330 $\pm$ 0.262 & 0.163 $\pm$ 0.163 & 0.387 $\pm$ 0.136 & 0.517 $\pm$ 0.017  \\
        \midrule
        \multirow{5}{*}{\shortstack{Convex\\Combination}}
        & \texttt{0.01}                & 0.063 $\pm$ 0.040 & 0.219 $\pm$ 0.121 & 0.296 $\pm$ 0.149 & 0.554 $\pm$ 0.056 & 0.509 $\pm$ 0.012  \\
        & \texttt{0.1}                 & \textbf{0.063 $\pm$ 0.039} & 0.219 $\pm$ 0.121 & 0.337 $\pm$ 0.167 & 0.571 $\pm$ 0.042 & 0.511 $\pm$ 0.013  \\
        & \texttt{0.5}                 & 0.065 $\pm$ 0.040 & 0.219 $\pm$ 0.122 & 0.380 $\pm$ 0.254 & 0.533 $\pm$ 0.059 & 0.512 $\pm$ 0.021  \\
        & \texttt{0.9}                 & 0.065 $\pm$ 0.041 & 0.224 $\pm$ 0.130 & 0.168 $\pm$ 0.186 & 0.405 $\pm$ 0.092 & 0.522 $\pm$ 0.022  \\
        & \texttt{0.99}                & 0.114 $\pm$ 0.184 & 0.248 $\pm$ 0.168 & 0.184 $\pm$ 0.202 & 0.338 $\pm$ 0.135 & 0.520 $\pm$ 0.011  \\
        % \cline{1-2}
        \midrule
        \multirow{3}{*}{\shortstack{Perturbing\\Gradients}}
         & \texttt{PCGrad}             & 0.063 $\pm$ 0.040 & 0.216 $\pm$ 0.123 & 0.125 $\pm$ 0.228 & 0.410 $\pm$ 0.140 & 0.512 $\pm$ 0.013  \\
         & \texttt{MGDA}               & 0.063 $\pm$ 0.040 & 0.214 $\pm$ 0.119 & 0.148 $\pm$ 0.226 & 0.393 $\pm$ 0.141 & 0.509 $\pm$ 0.012  \\
         & \texttt{DCGD}               & 0.096 $\pm$ 0.175 & 0.211 $\pm$ 0.122 & 0.209 $\pm$ 0.188 & 0.366 $\pm$ 0.112 & 0.523 $\pm$ 0.018  \\
         \midrule
         \multirow{2}{*}{Ours}
         & \texttt{$\kappa=0$}         & 0.066 $\pm$ 0.047 & \textbf{0.201 $\pm$ 0.128} & 0.102 $\pm$ 0.073 & \textbf{0.278 $\pm$ 0.071} & \textbf{0.504 $\pm$ 0.011}  \\
         & \texttt{$\kappa=1$}         & 0.068 $\pm$ 0.047 & 0.204 $\pm$ 0.133 & \textbf{0.100 $\pm$ 0.078} & 0.288 $\pm$ 0.077 & \textbf{0.504 $\pm$ 0.011}  \\
        \bottomrule
        \end{tabular}
    \end{sc}
    \end{small}
    
    % \vskip -0.1in
    \label{tab:result}
\end{table*}

%%%%%%%%%%%%%%%%%%%%%%%%%%%%%%%%%%%%%%%%%% SECTION: EXPERIMENTS_RESULTS %%%%%%%%%%%%%%%%%%%%%%%%%%%%%%%%%%%%%%%%%%
\section{Experiments and Results}
\label{sec:exp}
We validate our approach on three stochastic optimization problems: knapsack, budget allocation, and portfolio optimization.

%%%%%%%%%%%%%%%%%%%%% SUBSEC: EXP SETTINGS %%%%%%%%%%%%%%%%%%%%%
\subsection{Experimental Settings}
\label{subsec:exp-settings}
In this section, we begin by describing the setup for each of the three stochastic optimization problems.  
Next, we outline the baseline methods used for comparison.  
Then, we present the evaluation metrics and implementation details used in our experiments.

%%%%%%%%%%%%%%%%%%%%% paragraph: problems %%%%%%%%%%%%%%%%%%%%%
\paragraph{Problem Description}
We evaluate our method on three stochastic optimization problems.
For the knapsack problem, we consider both the weighted and unweighted variants.
For the budget allocation problem, we evaluate performance on both the original setting and a more challenging version augmented with fake targets.
Detailed descriptions of each problem setting are provided below.

\begin{itemize}
    \item \textbf{Knapsack Problem} \citep{mandi2020smart}:
    The objective is to select a subset of items that maximizes the total value without exceeding a weight capacity.
    We consider two variants: the unweighted knapsack, where all items have unit weight, and the problem is solvable in polynomial time;
    and the weighted knapsack, where each item has a given weight, making the problem NP-hard.
    We use a capacity constraint of $\{25,35,45\}$ for the unweighted knapsack and $\{30,90,150\}$ for the weighted knapsack.
    
    \textit{Prediction}:
    The model predicts the value of each item given its feature vector.
    
    \textit{Optimization}:
    Select the subset of items that maximizes the total value, satisfying the capacity constraint.

    \item \textbf{Budget Allocation} \cite{wilder2019melding}:
    The goal is to allocate a fixed advertising budget across websites to maximize the probability that a user clicks on at least one advertisement.
    To introduce additional difficulty, we create a variant by appending 500 randomly generated click-through rates (CTRs), referred to as fake targets, to the original CTRs.  

    \textit{Prediction}:
    Predict the CTRs of users for each website given its features.
    
    \textit{Optimization}:
    Given the CTRs prediction, select websites that maximize the expected number of users who click the website at least once.

    \item \textbf{Portfolio Optimization} \cite{markowitz2008portfolio}:
    The objective is to allocate the weights to stocks that maximize the risk-adjusted return.
    
    \textit{Prediction}:
    Predict the future returns of each stock based on the historical data.
    
    \textit{Optimization}:
    Choose weights for each stock that maximize the expected risk-adjusted return.
    
\end{itemize}

Further details on experiment settings such as datasets and optimization formulations are provided in Appendix \ref{appen:setting}.

%%%%%%%%%%%%%%%%%%%%% paragraph: baselines %%%%%%%%%%%%%%%%%%%%%
\paragraph{Baselines}
We compare our method against a range of baselines, grouped into three categories: naive approaches, convex combination methods, and gradient perturbation methods.
Visualization on updating gradients for each baseline is depicted in Figure \ref{fig:gradients}.
Each category is described below.
\begin{itemize}
    \item \textbf{Naive:}
        Two standard baselines, which are \texttt{PFL}, which trains solely using the prediction loss $\mathcal{L}_{pred}$, and \texttt{DFL}, which uses only the decision loss $\mathcal{L}_{dec}$.
        For DFL, we utilize differentiable solvers \cite{agrawal2019differentiable, wilder2019melding} to enable gradient flow through the optimization layer.
    \item \textbf{Convex Combination:}
        Baselines using the convex combination of $\mathcal{L}_{pred}$ and $\mathcal{L}_{dec}$.
        We combine the loss as $(1-\beta) \mathcal{L}_{pred} + \beta \mathcal{L}_{dec}$ where $\beta \in [0,1]$ controls the influence of decision loss.
        We use $\beta \in \{ \texttt{0.01}, \texttt{0.1}, \texttt{0.5}, \texttt{0.9}, \texttt{0.99} \}$ for our experiment.
    \item \textbf{Perturbing Gradients:}
        Although originally developed for general multi-objective optimization, these methods are conceptually aligned with our approach in that they adjust gradient directions to resolve conflicts across multiple objectives.
        We utilize \texttt{PCGrad}\cite{yu2020gradient}, \texttt{MGDA}\cite{desideri2012multiple}, \texttt{DCGD}\cite{hwang2024dual} as baselines.
\end{itemize}

%%%%%%%%%%%%%%%%%%%%% paragraph: metric %%%%%%%%%%%%%%%%%%%%%
\paragraph{Evaluation Metric}
We use the normalized test regret as our evaluation metric, where regret is defined as the difference between the decision loss achieved using the model’s prediction and the optimal objective value obtained using the ground-truth parameters $y$:
\begin{equation}
    \mathcal{R}(\hat{y},y) \vcentcolon= \mathcal{L}_{dec}(a^*(\hat{y}),y) - \mathcal{L}_{dec}(a^*(y),y).
\end{equation}
To normalize the regret, we compute the worst-case regret $\mathcal{R}_{worst}$ for each problem and divide the regret of each algorithm as  $\mathcal{R}_{test} / \mathcal{R}_{worst}$.
Intuitively, the normalized test regret reflects how far a model’s decision is from the best possible outcome in $[0,1]$, with lower values indicating better performance.
Note that evaluating normalized test regret is equivalent to comparing the decision loss.

We define the worst-case objective as the following.
For the knapsack problem, we assume that no items are selected, resulting in a total value of zero.
For the budget allocation problem, we derive the decision given the negated ground-truth data $-y$ and calculate its objective.
In the portfolio optimization problem, we allocate the entire investment to the asset with the lowest predicted return.

%%%%%%%%%%%%%%%%%%%%% paragraph: implementation %%%%%%%%%%%%%%%%%%%%%
\paragraph{Implementation Details}
All predictive models use a one-hidden-layer multilayer perceptron.
For the knapsack and budget allocation problems, we use 10 hidden nodes.
For the portfolio optimization problem, we use 500 hidden nodes.
All models are trained with the Adam optimizer at a learning rate of 0.001.
We fix the inflection point $c=50$ for our model.
Each experiment is run 10 times.

%%%%%%%%%%%%%%%%%%%%% subsec: Results %%%%%%%%%%%%%%%%%%%%%
\subsection{Results}
\label{subsec:results}

Table \ref{tab:result} reports the normalized test regret with standard error mean (SEM) for five experimental settings:
unweighted and weighted knapsack, budget allocation with 0 and 500 fake targets, and portfolio optimization.
We organize the baselines into four categories as in Section \ref{sec:exp} with our methods.

In both knapsack problems, we test for three different capacity constraints.
The naive baselines suffer from high regret and large SEM, showing low consistency.
The results of convex combination and perturbing gradients baselines imply that incorporating both gradients is beneficial in unweighted and weighted knapsack problems.

In both budget allocation problems, DFL outperforms PFL, highlighting the role of $\nabla \mathcal{L}_{dec}$.
As $\beta$ increases in the convex combination baselines, regret steadily decreases, confirming the emphasis on $\nabla \mathcal{L}_{dec}$ improves performance.
Gradient perturbation methods match this trend to some extent but remain less robust in the more challenging 500-fake scenario.
Our method consistently attains the best performance and maintains low SEM due to careful consideration of $\nabla \mathcal{L}_{dec}$.

The portfolio optimization problem differs from the previous problems, as accurate predictions are more directly related to high-quality decisions.
Unlike in other problems, PFL outperforms DFL in this setting.
Among convex combination baselines, smaller values of $\beta$ yield lower regret, reflecting the influence of $\nabla \mathcal{L}_{pred}$.
The gradient perturbation baselines also perform well by utilizing both gradient information.
Our method achieves the lowest regret and SEM, demonstrating both effectiveness and consistency.

Overall, our approach is the only method that consistently outperforms across all tasks, delivering the lowest regret in most problem settings and exhibiting strong reliability with low SEM.

%%%%%%%%%%%%%%%%%%%%%%%%%%%%%%%%%%%%%%%%%% SECTION: CONCLUSION %%%%%%%%%%%%%%%%%%%%%%%%%%%%%%%%%%%%%%%%%%
\section{Conclusion}
\label{sec:conclusion}

We have introduced a simple yet principled approach for DFL that leverages both prediction and decision loss gradients dynamically.
Unlike approaches that require constructing surrogate losses or learning additional parameters, our method can be plugged into any existing differentiable DFL algorithm with no extra training overhead.
A key advantage of our approach is its universality: by guiding existing DFL algorithms with prediction-guided perturbations, we enhance their stability and decision quality.
This plug-and-play nature makes it straightforward to integrate with new differentiable solvers and problem formulations, broadening the applicability of the DFL training paradigm in real-world settings.

Empirically, our method outperforms not only in normalized test regret but also in terms of stability across various stochastic optimization problems: unweighted and weighted knapsack, budget allocation, and portfolio optimization.

Theoretical analysis establishes that our method avoids conflicting gradients and converges to a Pareto stationary point under mild assumptions, offering insight into why it performs robustly across tasks.
Empirical results support this robustness: even when traditional PFL and DFL struggle, our method maintains low regret.

Despite these strengths, our work has two notable limitations.
First, the current formulation assumes access to exact decision loss gradients.
While this aligns with the classic DFL setting, much of the recent research has focused on surrogate losses; extending our method to those contexts remains an open direction.
Second, while our approach remains effective when either PFL or DFL underperforms, it is unclear how it behaves when both gradients lead to catastrophically poor results.
An edge case worth exploring further.
We leave these limitations for future research.

%%%%%%%%%%%%%%%%%%%%% NOTES %%%%%%%%%%%%%%%%%%%%%
% \begin{itemize}
%     \item limitations
% \end{itemize}
%%%%%%%%%%%%%%%

\bibliography{aaai2026}

%%%%%%%%%%%%%%%%%%%%%%%%%%%%%%%%%%%%%%%%%% SECTION: APPENDIX %%%%%%%%%%%%%%%%%%%%%%%%%%%%%%%%%%%%%%%%%%
% \input{ReproducibilityChecklist}

\newpage
\appendix
\onecolumn
%%%%%%%%%%%%%%%%%%%%%%%%%%%%%%%%%%%%%%%%%% SECTION: APPENDIX %%%%%%%%%%%%%%%%%%%%%%%%%%%%%%%%%%%%%%%%%%

%%%%%%%%%%%%%%%%%%%%%%%%%%%%%%%%%%%%%%%%%% SEC: NOTATIONS %%%%%%%%%%%%%%%%%%%%%%%%%%%%%%%%%%%%%%%%%%
% \section{Notations}
% \label{appen:notation}
% These are the notations used in the paper.

% %%%%%%%%%%%%%%%%%%%%% TABLE: notation %%%%%%%%%%%%%%%%%%%%%
% \begin{table*}[h]
% \caption{notation table}
% \begin{tabular}{cl}
%     \toprule
%     Notation & Description \\
%     \midrule
%     $\mathcal{D}$           & dataset \\
%     $\mathcal{M}_{\theta}$  & predictive model parameterized by $\theta$ \\
%     $x$                     & the first alphabet \\
%     $y$                     & the third alphabet \\
%     $\hat{y}$               & the third alphabet \\
%     $a^*(\hat{y})$          & the third alphabet \\
%     $\mathcal{L}_{pred}$    & prediction loss \\
%     $\mathcal{L}_{dec}$     & decision loss \\
%     $u_i$                   & unit gradient of $i$ \\
%     $\alpha$                & -             \\
%     $c$                     & decision loss \\
%     $\kappa$                & decision loss \\
%     $g$                     & decision loss \\
%     $m$                     & decision loss \\
%     $\phi$                     & decision loss \\
%     \bottomrule
% \end{tabular}
% \label{tab:notation}
% \end{table*}

%%%%%%%%%%%%%%%%%%%%%%%%%%%%%%%%%%%%%%%%%% SEC: PROOFS %%%%%%%%%%%%%%%%%%%%%%%%%%%%%%%%%%%%%%%%%%
\section{Proofs for Theorems}
\label{appen:proof}

\paragraph{Proposition 3.2}
\label{prop:appen-no-conf}
Our method never conflicts with $\nabla \mathcal{L}_{dec}$.
Moreover, when $\kappa=0$, our method does not conflict with both $\nabla \mathcal{L}_{pred}$ and $\nabla \mathcal{L}_{dec}$.

\begin{proof}
Since $0 < \alpha \leq 1$, the angle $ \angle \bigl( \nabla \mathcal{L}_{dec}, g \bigr) $ where $g$ is the merged gradient defined in Equation \ref{eq:g}, is at most $ \angle \bigl( \nabla \mathcal{L}_{dec}, u_{pred} + u_{dec} \bigr). $
Since $\phi_{pred, dec} \leq \pi$ and $\phi_{(pred+dec), dec} \leq \pi/2$, we have $\cos (\phi_{(pred+dec), dec} ) \geq 0$.
By Definition \ref{def:conflict}, our method \textit{never conflicts} with $\nabla \mathcal{L}_{dec}$.

When $\kappa=0$, the decay parameter $\alpha=1$, so the merged gradient $g$ always halves the angle of $\nabla \mathcal{L}_{pred}$ and $\nabla \mathcal{L}_{dec}$.
Since $0 \leq \phi_{pred, dec} \leq \pi$, it leads to $0 \leq \phi_{g, pred} \leq \frac{\pi}{2}$ and $0 \leq \phi_{g, dec} \leq \frac{\pi}{2}$.
Thus, by Definition \ref{def:conflict}, our method with $\kappa=0$ \textit{never conflicts} with both $\nabla \mathcal{L}_{pred}$ and $\nabla \mathcal{L}_{dec}$.
\end{proof}

%%%%%%%%%%%%%%%%%%%%% THM: Pareto-stationary %%%%%%%%%%%%%%%%%%%%%
\paragraph{Theorem 3.4}
\label{thm:appen-pareto}
% \begin{theorem}[Convergence to Pareto Stationary Point] 
Assume $\mathcal{L}_{pred}$ and  $\mathcal{L}_{dec}$ are differentiable with $M_i$-Lipschitz continuous with $M_i>0$.
Then, our method with $\kappa=0$ converges to a Pareto-stationary point for the step size $\eta_k$ satisfying $ \sum_{k=0}^\infty\eta_k=\infty$ and $\sum_{k=0}^\infty\eta_k^2<\infty$.
% \begin{gather}
%     \sum_{k=0}^\infty\eta_k=\infty,
%     \quad
%     \sum_{k=0}^\infty\eta_k^2<\infty
% \end{gather}
Moreover, the algorithm converges at a rate upper bounded by $\mathcal{O}(1/\sqrt{T})$.
% \end{theorem}

%%%%%%%%%%%%%%%%%%%%% PF: pareto stationarity %%%%%%%%%%%%%%%%%%%%%
\begin{proof}
At iterate $x_k$, define

\begin{align*}
 u_{pred}^k \vcentcolon= \frac{\nabla \mathcal L_{pred}(x_k)}{\|\nabla \mathcal L_{pred}(x_k)\|},
 \quad
 u_{dec}^k  \vcentcolon= \frac{\nabla \mathcal L_{dec}(x_k)}{\|\nabla \mathcal L_{dec}(x_k)\|},
\end{align*}

and
\begin{align*}
 d_k \vcentcolon= \frac{u_{pred}^k + u_{dec}^k}{\|u_{pred}^k + u_{dec}^k\|},
 \quad\text{if }u_{pred}^k+u_{dec}^k\neq0,
\end{align*}

else stop.  Set the geometric mean of $\nabla \mathcal{L}_{pred}$ and $\nabla \mathcal{L}_{dec}$ as
\begin{align*}
 m_k = \sqrt{\|\nabla \mathcal L_{pred}(x_k)\| \cdot \|\nabla \mathcal L_{dec}(x_k)\|},
\end{align*}

and assume $m_k\le A$. Define the aggregate objective
\begin{align*}
 \mathcal L(x) \vcentcolon= \mathcal L_{pred}(x) + \mathcal L_{dec}(x),
\end{align*}

where $\mathcal{L}(x)$ is bounded below. Using the differentiability and Lipschitz continuity, for $x_{k+1} \vcentcolon= x_k - \eta_k m_k d_k$ we have
\begin{align*}
 \mathcal{L}_i(x_{k+1}) - \mathcal{L}_i(x_k)
 &\leq  \langle \nabla \mathcal L_{i}(x_k), x_{k+1}-x_k \rangle + \tfrac{M_{i}}{2} \| x_{k+1} - x_k \|^2 \\
 &\leq -\eta_k\,m_k\,\langle \nabla \mathcal L_{i}(x_k), d_k \rangle + \tfrac{M_{i}}{2}\,\eta_k^2\,m_k^2,
\end{align*}

for $i \in \{pred, dec \}$. Summing and letting $M \vcentcolon= M_{pred}+M_{dec}$ yields
\begin{align} \label{eq:tele}
 \mathcal L(x_{k+1}) - \mathcal L(x_k) \leq
 - \eta_k\,m_k\,\psi_k + \tfrac{M}{2}\,\eta_k^2\,m_k^2,
\end{align}

where
\begin{align*}
 \psi_k
 &= \langle \nabla \mathcal L_{pred}(x_k), d_k \rangle + \langle \nabla \mathcal L_{dec}(x_k), d_k  \rangle \\
 &= \|\nabla \mathcal L_{pred}(x_k)\| \cdot \|d_k\| \cos\frac{\phi_k}{2}
 + \|\nabla \mathcal L_{dec}(x_k)\| \cdot \|d_k\| \cos\frac{\phi_k}{2} \\
 &= \bigl(\|\nabla \mathcal L_{pred}(x_k)\| + \|\nabla \mathcal L_{dec}(x_k)\|\bigr) \cos\frac{\phi_k}{2} \\
 % &> 0,
\end{align*}

with $\phi_k=\angle(u_{pred}^k,u_{dec}^k)$. Using telescoping sum in Equation \ref{eq:tele}, we have
\begin{align} \label{eq:bounded}
 \mathcal L(x_0) - \mathcal L(x^*) 
 \geq \sum_{k=0}^{\infty}\bigl( \eta_k\,m_k\,\psi_k - \tfrac{M}{2}\,\eta_k^2\,m_k^2 \bigr).
 \end{align}

Since $\sum\eta_k^2<\infty$ and $m_k\le A$, it follows
\begin{align} \label{eq:A}
\sum_{k=0}^{\infty} \eta_k^2 m_k^2 \leq A^2 \sum_{k=0}^{\infty} \eta_k^2 < \infty.
\end{align}

Using Equation \ref{eq:bounded}, \ref{eq:A} and that $\mathcal{L}$ is bounded below, we have
\begin{align} \label{eq:bounded2}
\sum_{k=0}^{\infty} \eta_k m_k\psi_k
&\leq \mathcal L(x_0) - \mathcal L(x^*) + \tfrac{M}{2}\sum_{k=0}^{\infty}\eta_k^2 m_k^2 \nonumber \\
&\leq \mathcal L(x_0) - \mathcal L(x^*) + \tfrac{MA^2}{2}\sum_{k=0}^{\infty}\eta_k^2 \nonumber \\
&< \infty.
\end{align}

From Equation \ref{eq:bounded2} and $\sum_{k=0}^{\infty}\eta_k = \infty$, it follows
\begin{align*}
\liminf_{k\to\infty}(m_k\psi_k)=0.
\end{align*}

Thus, along some subsequence $\{ x_k \}$, either $\|\nabla \mathcal L_{pred}(x_k)\| \to 0$, or $\|\nabla \mathcal L_{dec}(x_k)\| \to 0$, or $\cos\frac{\phi_k}{2} \to 0$, each implying \textit{Pareto stationarity}.

%%%%%%%%%%%%%%%%%%%%% PF: convergence rate %%%%%%%%%%%%%%%%%%%%%
Now, choose the step size $\eta_k$ as
\begin{align*}
\eta_k = \frac{\eta_0}{(k+1)^\alpha}, \quad \text{with }\alpha\in(\tfrac12,1),
\end{align*}

satisfying the step size assumption. For step $T$, define
\begin{align*}
 P_T
 = \sum_{k=0}^{T-1}\eta_k ,
 \quad
 Q_T
 = \sum_{k=0}^{T-1}\eta_k^2 .
\end{align*}

Then for $\alpha \in (\frac{1}{2},1)$, 
\begin{align*}
P_T &= \sum_{k=0}^{T-1}\frac{\eta_0}{(k+1)^\alpha} \\
&\geq
\eta_0 \int_{1}^{T} x^{-\alpha}\,dx \\
&=
\frac{\eta_0}{1-\alpha}(T^{1-\alpha}-1)
\end{align*}

and
\begin{align*}
Q_T &=\sum_{k=0}^{T-1}\frac{\eta_0^2}{(k+1)^{2\alpha}} \\
&\leq
\eta_0^2 \bigl( 1 + \int_{1}^{\infty} x^{-2\alpha}\,dx \bigr) \\
&=
\eta_0^2 \bigl( 1 + \frac{1}{2\alpha-1} \bigr).
\end{align*}

% Using the telescoping sum we have,
% \begin{align}
% \label{eq1}
%  \sum_{k=0}^{T-1}\eta_k m_k\psi_k\le \mathcal L(x_0)-\mathcal L(x^*) + \tfrac{M}{2}A^2Q_T.
% \end{align}

Note that
\begin{align}
\label{eq:note}
\sum_{k=0}^{T-1}\eta_k\,m_k\,\psi_k
& \ge \bigl(\min_{0\le k< T}(m_k\psi_k)\bigr)\,
\sum_{k=0}^{T-1}\eta_k \nonumber \\
&= \bigl(\min_{0\le k< T}(m_k\psi_k)\bigr)\,P_T.
\end{align}

Then, combining Equation \ref{eq:bounded2} and \ref{eq:note} we have
\begin{align*}
\min_{0\le k< T}(m_k\psi_k)\;P_T
\;\le\;
\mathcal L(x_0) - \mathcal{L}(x^*) + \frac{MA^2}{2}\,Q_T .
\end{align*}

Finally,
\begin{align*}
 \min_{0\le k< T}(m_k\psi_k)
 &\;\le\; \frac{\mathcal L(x_0) - \mathcal{L}(x^*) + \tfrac{M A^2}{2}Q_T}{P_T} \\
 &= \mathcal{O} \bigl( T^{-(1-\alpha)} \bigr) \\
 &< \mathcal{O} \bigl( \frac{1}{\sqrt{T}} \bigr) .
\end{align*}

Since $m_k\psi_k = \langle \nabla \mathcal{L}(x_k), m_kd_k \rangle$ measures how far we are from satisfying the Pareto stationary condition, we have the proof.
\end{proof}

%%%%%%%%%%%%%%%%%%%%%%%%%%%%%%%%%%%%%%%%%% SEC: SETTINGS %%%%%%%%%%%%%%%%%%%%%%%%%%%%%%%%%%%%%%%%%%
\section{Experiment Settings Details}
\label{appen:setting}

%%%%%%%%%%%%%%%%%%%%% subsec: code link %%%%%%%%%%%%%%%%%%%%%
\subsection{Codes for Paper}
The codes for the experiments are available at the following link.
\begin{links}
  \link{Code}{https://anonymous.4open.science/r/twilight-1B7C}
\end{links}

%%%%%%%%%%%%%%%%%%%%% subsec: knapsack %%%%%%%%%%%%%%%%%%%%%
\subsection{Details in Knapsack Problem}

For the knapsack problem, we use the data from \citet{mandi2020smart}.
The objective of a knapsack problem is to select a subset of items that maximizes total value without exceeding a capacity limit.
We consider the two variants of the knapsack problem: unweighted and weighted knapsack.
For total number of $N=48$ items, each item $i$ has the features $x_i \in \mathbb{R}^{8}$ and the corresponding value $v_i \in \mathbb{R}$ and the weight $w_i \in \mathbb{R}$.
In the unweighted knapsack problem, we assume the weights are fixed to 1.
For the weighted knapsack problem, we use the given weights from the data, sampled from $\{3,5,7\}$.
We conduct experiments on three different capacity constraints $c \in \mathbb{R}$ for each problem: $\{25, 35, 45\}$ for the unweighted knapsack, and $\{30, 90, 150\}$ for the weighted knapsack.

Given the item features and weights, we first predict the value $v_i$.
Then, we choose the items to select using a binary vector $a \in \mathbb{R}^{N}$ that maximizes the sum of the values of items under the capacity constraint:
\begin{equation}
\begin{aligned}
    \max_{a} \quad & \sum_{i=1}^{N} v_i a_i \notag \\
    \text{s.t.} \quad & \sum_{i=1}^{N} w_i a_i \leq c \\
        \quad & a\in \{0, 1\}^N
\end{aligned}
\end{equation}

%%%%%%%%%%%%%%%%%%%%% subsec: budget %%%%%%%%%%%%%%%%%%%%%
\subsection{Details in Budget Allocation Problem}

% The objective of the submodular optimization problem \cite{wilder2019melding} is to choose $B=2$ websites to advertise based on click-through rates (CTRs) of $U=10$ users among $W=5$ websites.
% We generate the website features $x_w \in \mathbb{R}^{U}$ for each website $w$ by multiplying a random matrix $A \in \mathbb{R}^{U \times U}$ to the CTRs $y_w \in \mathbb{R}^{U}$.
% We conduct experiments with $\{0, 5, 50, 500\}$ fake targets to add difficulty by concatenating the fake target size of random CTRs to the original CTRs.
% Given the website features $x_w$ for each website $w$, we first predict the CTRs $\hat{y}_w$.
% Then, we choose $B$ websites to advertise using a binary vector $a \in \mathbb{R}^{W}$ that maximizes the expected number of users that click on the advertisement at least once:

% \begin{equation}
% \begin{aligned}
%     \max_{a} \quad & \sum_{u=1}^{U}(1 - \prod_{w=1}^{W} (1 - a_w \cdot \hat{y}_{wu})) \notag \\
%     \text{s.t.} \quad & a\in \{0, 1\}^W
% \end{aligned}
% \end{equation}

% %%%%%%%%%%%%%%%%%%%%% subsec: budget %%%%%%%%%%%%%%%%%%%%%
% \subsection{Experimental Setup for Budget Allocation Task}

Following the setup introduced by \citet{wilder2019melding}, we address a submodular optimization task in which the goal is to select \( B = 2 \) websites from a pool of \( W = 5 \) websites to advertise to \( U = 10 \) users, using predicted click-through rates (CTRs).
To simulate website features, we construct a feature vector \( x_w \in \mathbb{R}^U \) for each website \( w \) by applying a random linear transformation. Specifically, we multiply each website's true CTR vector \( y_w \in \mathbb{R}^U \) with a randomly generated matrix \( A \in \mathbb{R}^{U \times U} \). 

To introduce varying levels of complexity into the task, we consider two variants for the budget allocation problem that augment the CTR data with randomly generated fake targets.
These fake targets are appended to the original CTRs, increasing the input dimensionality and introducing noise into the prediction process.
We consider $\{0,500\}$ fake targets.

Given the generated features \( x_w \), the model predicts estimated CTRs \( \hat{y}_w \) for each website.
The selection of advertising sites is represented by a binary decision vector \( a \in \mathbb{R}^W \), where each entry indicates whether a website is selected.
The optimization objective is to maximize the expected number of users who click on at least one of the selected websites:

\begin{equation}
\begin{aligned}
    \max_{a} \quad & \sum_{u=1}^{U}\left(1 - \prod_{w=1}^{W} \left(1 - a_w \cdot \hat{y}_{wu}\right)\right) \notag \\
    \text{s.t.} \quad & a \in \{0, 1\}^W.
\end{aligned}
\end{equation}

%%%%%%%%%%%%%%%%%%%%% subsec: portfolio %%%%%%%%%%%%%%%%%%%%%
\subsection{Details in Portfolio Optimization Problem}

\begin{links}
\link{Data}{https://mba.tuck.dartmouth.edu/pages/faculty/ken.french/data_library.html}
\end{links}

For the portfolio optimization problem, we use the $N=49$ industry portfolios data from Kenneth R. French's homepage.
The link to the data is given above.
Given the historical data, we predict the future returns of each industry data $y \in \mathbb{R}^{N}$.
Then, we choose the portfolio weight $a \in \mathbb{R}^{N}$ that maximizes the expected risk-adjusted return given the covariance matrix $\Sigma \in \mathbb{R}^{N \times N}$ and the risk aversion coefficient $\lambda=1$:

\begin{equation}
\begin{aligned}
    \max_{a} \quad & y^\top a - \lambda \cdot a^\top \Sigma a \notag \\
    \text{s.t. } \quad & \sum_{i=1}^{N} a_i = 1.
\end{aligned}
\end{equation}
% where we set $\lambda=1$.

\end{document}